\documentclass[11pt]{article}
\usepackage[letterpaper,margin=1in]{geometry}
\usepackage[T1]{fontenc}
\usepackage{lmodern}
\usepackage{microtype}
\usepackage[hyphens]{url}
\usepackage{graphicx}
\usepackage{float}
\usepackage{booktabs}
\usepackage{amsmath}
\usepackage{amssymb}
\usepackage{multirow}
\usepackage[table]{xcolor}
\usepackage{placeins}
\usepackage[authoryear,round]{natbib}
\definecolor{linkblue}{RGB}{0,70,140}
\usepackage[colorlinks=true,allcolors=linkblue]{hyperref}

\title{Concept Labels Are Not Enough: Rethinking Concept Bottleneck Models through Representation Integrity}
\author{
  \makebox[\dimexpr\textwidth-2\tabcolsep\relax][c]{%
    \begin{minipage}[t]{0.48\textwidth}
      \centering
      \textbf{Gaoxiang Huang} \\
      Deep Interdisciplinary Intelligence Lab \\
      HKUST(GZ) \\
      Guangzhou, China \\
      \footnotesize\texttt{ghuang991@connect.hkust-gz.edu.cn}
    \end{minipage}%
    \hfill
    \begin{minipage}[t]{0.48\textwidth}
      \centering
      \textbf{Songning Lai, Yutao Yue}\thanks{Corresponding author: Prof. Yutao Yue} \\
      Deep Interdisciplinary Intelligence Lab \\
      HKUST(GZ) \\
      Guangzhou, China \\
      \footnotesize\texttt{\{songninglai, yutaoyue\}@hkust-gz.edu.cn}
    \end{minipage}%
  }
}
\date{}

\begin{document}
\maketitle

\begin{abstract}
Although deep neural networks achieve strong predictive performance, their internal reasoning often remains difficult to inspect and control. Concept Bottleneck Models (CBMs) address this opacity by factoring predictions through human-understandable concepts, thereby enabling concept-level inspection and intervention. However, CBMs remain vulnerable to concept shift and information leakage, while existing evaluations neither reveal how the internal features supporting each concept are organized nor identify the representation-level deficiency associated with these failures. We argue that this missing property is concept integrity: concept support should form a semantically coherent and non-fragmented functional group. To characterize this property, we propose group coherence (GC) and concept coverage (CC) as integrity components and aggregate them into the concept integrity score (CIS). We further introduce Concept Integrity Regularization (CIR) to encourage coherent and separated groups of concept-supporting filters. Our latent decoupled concept bottleneck model (LDCBM) applies CIR without requiring region annotations. Across three datasets, CIS rankings differ from rankings by concept and task accuracy, supporting concept integrity as a complementary representation-level criterion. Moreover, with only 10\% of concept labels, LDCBM retains approximately 93\% of its full-label task performance. Background-masking and concept-intervention stress tests further associate stronger integrity profiles with lower sensitivity to nuisance context and more responsive concept correction.
\end{abstract}

\section{Introduction}
\label{sec:intro}
Concept Bottleneck Models (CBMs) make predictions through an intermediate layer of human-defined concepts: an input is first mapped to concepts, and the task label is then predicted from those concepts~\cite{kohConceptBottleneckModels2020a}. This factorization exposes a high-level interface through which users can inspect model outputs, identify concept errors, and intervene by correcting predicted concepts. These properties have made CBMs a prominent family of interpretable-by-design models and have motivated a rapidly expanding set of architectural and training variants~\cite{knabWhatsBottleSurvey2026}.

The concept interface is informative, however, only when a predicted concept is supported by the intended evidence. Prior studies show that CBMs can satisfy output-level objectives while learning concept predictors that do not correspond to semantically meaningful input evidence~\cite{margeloiuConceptBottleneckLearn2021}, encode unintended task information in concept activations~\cite{havasiAddressingLeakageConcept2022}, or respond to features outside a concept's spatial or semantic locality~\cite{huangConceptTrustworthinessConcept2024a,ramanConceptBottleneckRespect2025}. Intervention effectiveness likewise depends on which concepts are corrected and under what conditions~\cite{shinCloserLookIntervention2023}. More recently, high predictive accuracy has been shown to persist even with irrelevant concepts when a linear pipeline can bypass the intended bottleneck~\cite{tapliRethinkingConceptBottleneck2026}. Together, these findings indicate that CBMs still face several important limitations. First, concept shift can arise when a model predicts multiple local concepts, such as part-level attributes, from a global image feature. This coarse input-to-concept mapping weakens the reliability of concept-based explanations. Second, information leakage occurs when concept activations contain unintended information that is used for downstream prediction. Crucially, this unintended information is not necessarily visible to humans.

\begin{figure}
    \centering
    \includegraphics[scale = 0.6]{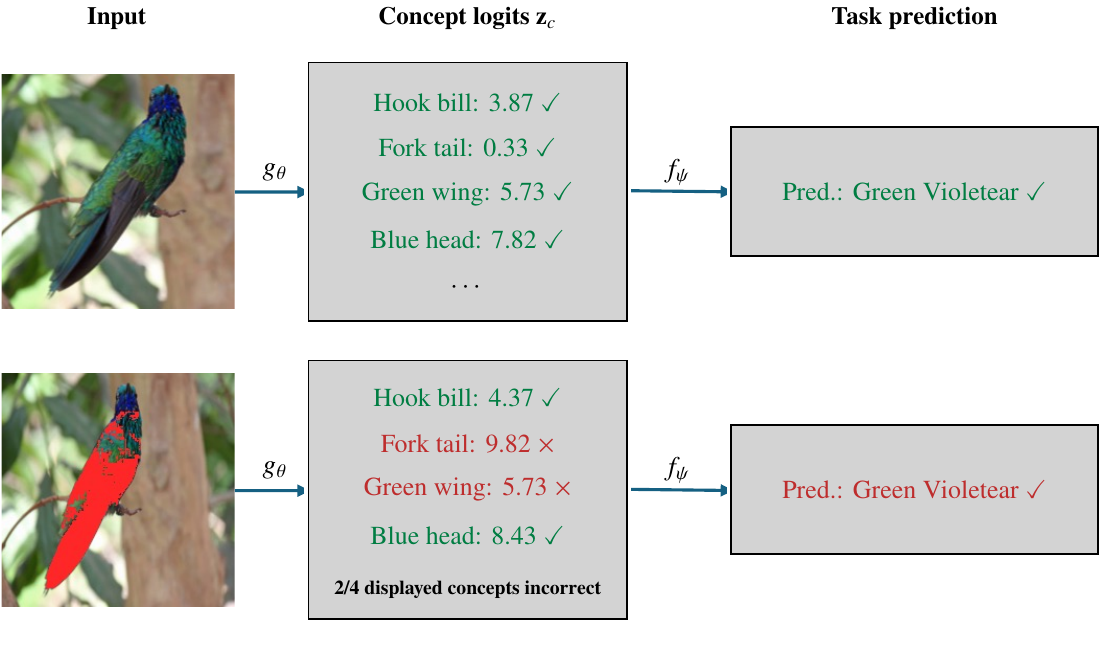}
    \caption{Correct task prediction can conceal failures in the concept pathway. After masking concept-relevant regions of the bird, the logits of several affected concepts increase rather than decrease, illustrating concept-shift behavior in the input-to-concept pathway; two of the four displayed concept predictions also become incorrect. Nevertheless, the CBM retains the correct Green Violetear classification, revealing a task–concept inconsistency that is consistent with unintended task-relevant information remaining in the soft concept representation. Values denote concept logits; $\checkmark$ and $\times$ indicate correct and incorrect concept predictions, respectively.}
    \label{fig:concept_integrity}
\end{figure}

Existing explanations often attribute these failures to limited concept annotation or to the information loss induced by a low-dimensional bottleneck. Inspired by the notion of neuronal assemblies, where a function is supported by an organized population rather than a single isolated unit~\cite{holtmaatNeuronalAssemblyFormation2016}, we argue that another representation-level factor is \emph{concept integrity}. A concept should not be treated as a single activation with a high response. Instead, it should be supported by a group of representations that are semantically consistent with the concept, sufficiently cover its visual evidence, and avoid unrelated information. When concept representations entangle low-level nuisance factors or label shortcuts, the downstream classifier can exploit non-conceptual information, leading to information leakage. The same unstable cues may break under distribution shift, resulting in concept shift and degraded generalization. Thus, the key distinction is not only whether a concept is labeled correctly, but also whether the representation supporting that concept is intact, as illustrated in Figure~\ref{fig:concept_integrity}.

To examine this view, we decompose concept integrity into two complementary properties. \emph{Group coherence} asks whether filters grouped together consistently support one dominant concept. \emph{Concept coverage} asks whether the filters relevant to a concept are concentrated in its primary group rather than fragmented across the representation. Group coherence (GC) and concept coverage (CC) operationalize these properties. Their geometric mean, the concept integrity score (CIS), is high only when both conditions hold. These metrics complement concept and task accuracy by asking how the evidence behind a concept prediction is organized.

We further introduce Concept Integrity Regularization (CIR) as a controlled structural intervention, which promotes high within-group and low cross-group activation similarity, allowing us to examine whether the diagnosed representation structure can be shifted without materially changing predictive performance. Our latent decoupled concept bottleneck model (LDCBM) applies CIR to standard CBM and serves as a concrete model for testing whether encouraging concept integrity changes representation organization and downstream behavior.

\begin{figure}[t]
\centering
\includegraphics[width=0.92\textwidth]{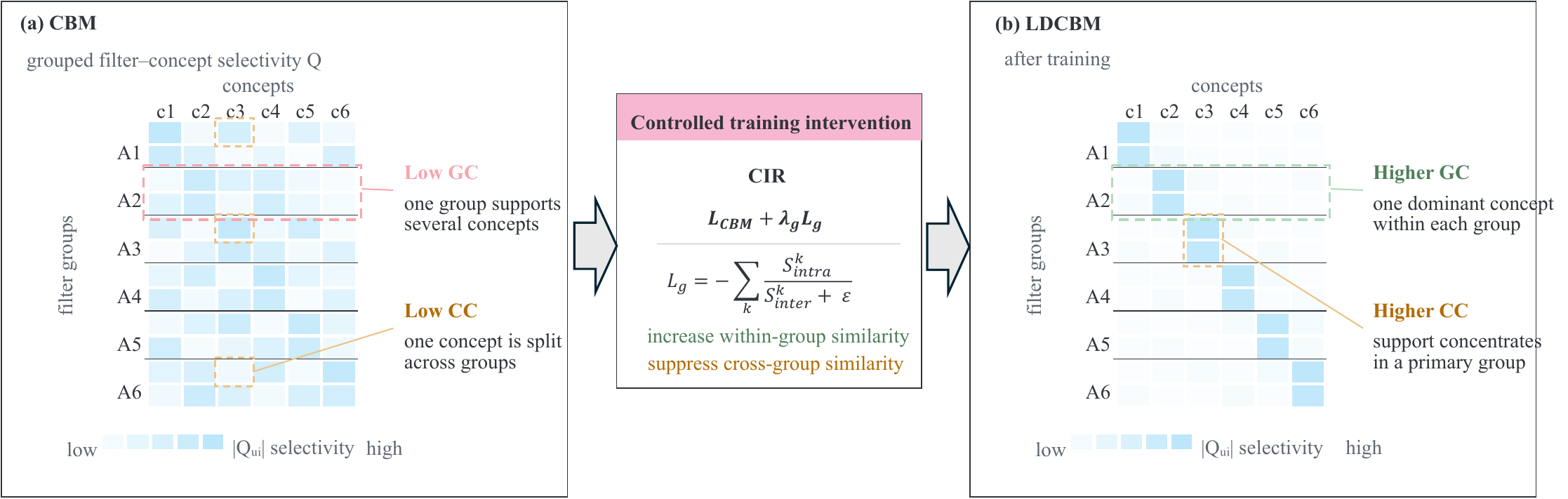}
\caption{Overview of the concept-integrity framework. We group filters from a standard CBM, measure semantic coherence and non-fragmentation with GC, CC and use CIR to encourage the corresponding representation structure. We stress-test the resulting integrity profiles through background masking, concept intervention, and scarce concept labels.}
\label{fig:framework}
\end{figure}

Figure~\ref{fig:framework} separates output-level concept supervision, representation-level concept integrity, and CIR as a way to encourage intact concept support.

We evaluate this view across three concept-annotated datasets and multiple CBM variants. GC, CC, and CIS capture information that is not reducible to concept or task accuracy: models with similar predictive performance can differ substantially in how their concept-supporting features are organized. Background-masking and concept-intervention experiments associate stronger integrity profiles with lower sensitivity to nuisance context and greater responsiveness to concept correction. 

Beyond these behavioral evaluations, we explore a natural extension: whether cleaner and more stable concept support also allows models to use limited concept annotations more effectively. Results under concept-label scarcity suggest that encouraging concept integrity can help preserve the concept--task pathway even when semantic supervision is limited.

Our contributions are threefold:
\begin{itemize}
    \item We decompose concept-pure support into semantic focus and evidence concentration, quantified by GC, CC, and CIS.
    \item We use CIR as a controlled structural intervention to test whether filter-group organization can be shifted while preserving predictive performance.
    \item We identify a hidden failure mode in CBMs: concept accuracy does not guarantee that concept predictions are supported by concept-integrity internal evidence.
\end{itemize}

\section{Related Work}

\paragraph{Concept bottlenecks and richer concept representations.}
The original CBM formulation routes predictions through supervised, human-interpretable concept values and supports direct intervention on those values~\cite{kohConceptBottleneckModels2020a}. Concept Embedding Models (CEMs) replace scalar concept values with higher-dimensional concept-specific embeddings to improve task utility and intervention behavior under incomplete supervision~\cite{zarlengaConceptEmbeddingModels2022a}. Subsequent work has diversified the input, concept, output, and training modules of CBMs, creating a broad design space with different assumptions about concept semantics, grounding, and completeness~\cite{knabWhatsBottleSurvey2026}. Our work is complementary to richer bottleneck designs: rather than increasing the expressiveness of a concept output, we examine how the internal filters supporting that output are organized.

\paragraph{Concept reliability, leakage, and intervention.}
Several studies question whether a transparent bottleneck is sufficient for faithful concept-based reasoning. Learned concepts may lack meaningful correspondence to input evidence~\cite{margeloiuConceptBottleneckLearn2021}, while soft concept activations may carry unintended task information that weakens interpretability and intervention~\cite{havasiAddressingLeakageConcept2022}. Spatial trustworthiness and locality analyses further show that a concept can depend on irrelevant image regions or features outside its intended support~\cite{huangConceptTrustworthinessConcept2024a,ramanConceptBottleneckRespect2025}. Other work studies intervention selection and granularity~\cite{shinCloserLookIntervention2023}, adversarial perturbations of concept-based models~\cite{sinhaUnderstandingEnhancingRobustness2023}, and structural cases in which high accuracy does not imply that the concept layer is functionally meaningful~\cite{tapliRethinkingConceptBottleneck2026}. We do not treat these failures as consequences of a single mechanism; instead, we ask whether group-level concept integrity supplies complementary information about the representation behind concept predictions.

\paragraph{Concept-representation evaluation.}
Zarlenga et al.~\cite{zarlengaRobustMetricsConcept2023} directly study impurities in concept representations and introduce purity metrics for concept learning and disentanglement models. Their work establishes that supervision alone need not produce semantically clean concept representations and evaluates consequences for robustness and intervention. We share the premise that output accuracy is insufficient. Our operational focus is the organization of concept-supporting CNN filters: GC measures semantic coherence within groups, CC measures whether support is fragmented across groups, and CIS requires both properties. We use \emph{integrity} to name this joint group-level condition, not to relabel prior purity metrics as a new general problem.

\section{Method}
\label{sec:method_rewrite}

We investigate concept integrity as a representation-level factor shared by three practical CBM challenges. Concept supervision constrains predicted bottleneck values, but not the internal features supporting each concept. Those features may mix semantic targets or scatter the support for one concept across several groups. We therefore ask whether concept evidence forms a semantically coherent and non-fragmented functional group. Three complementary metrics diagnose these properties, and CIR shapes the same shared feature representation. Complete assignments, thresholds, and aggregation rules are provided in Appendix~\ref{app:complete_integrity_definitions}.

\subsection{Problem Formulation and Shared Representation}
\label{sec:prelim_rewrite}

\paragraph{Concept bottleneck models.}
Let \(\mathcal{D}=\{(x^{(n)},\mathbf{c}^{(n)},y^{(n)})\}_{n=1}^{N}\) be a concept-annotated dataset, where \(x^{(n)}\in\mathcal{X}\), \(\mathbf{c}^{(n)}\in\{0,1\}^{M}\), and \(y^{(n)}\in\mathcal{Y}\). A CBM consists of a concept predictor \(g_{\theta}:\mathcal{X}\rightarrow[0,1]^M\) and a label predictor \(f_{\psi}:[0,1]^M\rightarrow\mathcal{Y}\)~\cite{kohConceptBottleneckModels2020a}. Its two-stage prediction, shown in Eq.~\ref{eq:rewrite_cbm_forward}, is
\begin{equation}
\hat{\mathbf{c}}^{(n)}
=g_{\theta}(x^{(n)}),
\qquad
\hat{y}^{(n)}
=f_{\psi}(\hat{\mathbf{c}}^{(n)}).
\label{eq:rewrite_cbm_forward}
\end{equation}
For joint training, the standard supervised objective is
\begin{equation}
\begin{aligned}
\mathcal{L}_{\mathrm{CBM}}(\theta,\psi)
&=
\frac{1}{N}\sum_{n=1}^{N}
\ell_y(\hat{y}^{(n)},y^{(n)})\\
&\quad+
\frac{\lambda_c}{NM}
\sum_{n=1}^{N}\sum_{i=1}^{M}
\ell_c(\hat{\mathbf{c}}^{(n)}_i,\mathbf{c}^{(n)}_i),
\end{aligned}
\label{eq:rewrite_cbm_objective}
\end{equation}
where \(\ell_y\) is the task loss, \(\ell_c\) is the per-concept prediction loss, and \(\lambda_c\) controls the strength of concept supervision.

\paragraph{Concept-labelled versus intact.}
Equation~\ref{eq:rewrite_cbm_objective} constrains the bottleneck output, but not how information is organized within the features that support each concept. A concept can therefore be predicted accurately while its support remains mixed or fragmented. We use concept-labelled for output-level supervision and intact for a representation whose concept-relevant filters form a semantically coherent and non-fragmented functional group. Integrity does not by itself establish that these filters are grounded in the intended visual evidence.

\paragraph{A unified activation representation.}
All diagnostics and CIR operate on the same pooled filter activations. For filter \(u\) in a selected convolutional layer, let
\begin{equation}
\mathbf{A}_{n,u}
=
\operatorname{GAP}\!\left(z_u(x^{(n)})\right),
\qquad
\mathbf{A}\in\mathbb{R}^{N\times|\Omega|}.
\label{eq:rewrite_activation_matrix}
\end{equation}
With the concept matrix \(\mathbf{C}\in\{0,1\}^{N\times M}\), the activation matrix in Eq.~\ref{eq:rewrite_activation_matrix} induces the two views in Eq.~\ref{eq:rewrite_selectivity}:
\begin{equation}
\begin{aligned}
Q_{u,i}
&=
\left|\operatorname{corr}(\mathbf{A}_{:,u},\mathbf{C}_{:,i})\right|,\\
S_{u,v}
&=
\operatorname{corr}(\mathbf{A}_{:,u},\mathbf{A}_{:,v})+1.
\end{aligned}
\label{eq:rewrite_selectivity}
\end{equation}
The filter--concept selectivity matrix \(\mathbf{Q}\) supports diagnosis, whereas the filter-similarity matrix \(\mathbf{S}\) supports grouping and CIR. Following the group-based organization of CNN filters~\cite{shenInterpretableCompositionalConvolutional2021}, spectral clustering on \(\mathbf{S}\) partitions filters into disjoint groups \(\mathcal{A}=\{\mathcal{A}_1,\ldots,\mathcal{A}_K\}\). Thus, the metrics and CIR diagnose and shape the same activation structure rather than introducing separate representations.

\subsection{Three Complementary Integrity Questions}
\label{sec:rewrite_metrics}

The three metrics are organized by the representation property they diagnose. Group coherence asks whether a learned group is semantically coherent. Concept coverage asks whether the filters relevant to a concept are concentrated in its primary group rather than fragmented across groups. The concept integrity score asks whether coherence and non-fragmentation hold simultaneously. All three are computed from \(\mathbf{Q}\) and the same partition \(\mathcal{A}\).

\paragraph{Question 1: Is each group coherent for one concept?}
Each group \(\mathcal{A}_k\) is associated with its most selective concept \(\phi(k)\). Group coherence is defined in Eq.~\ref{eq:rewrite_gc}; it compares every filter's selectivity to this group concept against that filter's strongest concept selectivity:
\begin{equation}
\operatorname{GC}
=
\frac{1}{K}\sum_{k=1}^{K}
\frac{1}{|\mathcal{A}_k|}
\sum_{u\in\mathcal{A}_k}
\frac{Q_{u,\phi(k)}}{\max_j Q_{u,j}+\epsilon}.
\label{eq:rewrite_gc}
\end{equation}
High GC indicates that filters grouped together consistently support the same dominant concept. It therefore diagnoses semantic mixing within groups.

\paragraph{Question 2: Is concept support non-fragmented?}
High group coherence can still arise when a concept's relevant filters are fragmented across multiple groups. For concept \(i\), let \(\mathcal{R}_i\) be its adaptively selected relevant-filter set and let \(\psi(i)\) be its most selective group. Concept coverage, defined in Eq.~\ref{eq:rewrite_cc}, measures the fraction of \(\mathcal{R}_i\) captured by that primary group:
\begin{equation}
\operatorname{CC}
=
\frac{1}{M}\sum_{i=1}^{M}
\frac{|\mathcal{R}_i\cap\mathcal{A}_{\psi(i)}|}
{|\mathcal{R}_i|+\epsilon}.
\label{eq:rewrite_cc}
\end{equation}
We use this concentration as the operational definition of non-fragmented support: high CC indicates that the filters relevant to a concept are concentrated in one primary group rather than dispersed across the representation. Appendix~\ref{app:complete_integrity_definitions} gives the adaptive threshold and group-selection rules.

\paragraph{Question 3: Are coherence and non-fragmentation jointly high?}
GC and CC capture complementary requirements. A group may be highly selective to one concept while containing only a fragmented subset of that concept's evidence; conversely, a densely populated group may mix evidence from unrelated concepts. We therefore combine semantic coherence and concept coverage using the geometric mean in Eq.~\ref{eq:rewrite_cis}:
\begin{equation}
\operatorname{CIS}
=
\sqrt{\operatorname{GC}\cdot\operatorname{CC}}.
\label{eq:rewrite_cis}
\end{equation}
CIS is high only when both properties are high, preventing one component from compensating for a weak value of the other. We use CIS as the operational measure of concept integrity: a high score requires concept-supporting features to be semantically coherent and non-fragmented. We then test whether this integrity profile is associated with behavior under masking, intervention, and scarce concept labels.

\subsection{Encouraging Concept Integrity with CIR}
\label{sec:rewrite_group_loss}

The metrics diagnose a learned representation but do not change it. CIR therefore operates on the filter-similarity view \(\mathbf{S}\). For each group \(\mathcal{A}_k\), we define the within-group and cross-group similarities as
\begin{equation}
\begin{aligned}
S_k^{\mathrm{intra}}
&=
\sum_{u,v\in\mathcal{A}_k}s_{uv},\\
S_k^{\mathrm{inter}}
&=
\sum_{\substack{u\in\mathcal{A}_k\\
v\in\Omega\setminus\mathcal{A}_k}}s_{uv}.
\end{aligned}
\label{eq:rewrite_intra_inter}
\end{equation}
Using the within-group and cross-group similarities in Eq.~\ref{eq:rewrite_intra_inter}, the CIR objective in Eq.~\ref{eq:rewrite_group_loss} is
\begin{equation}
\mathcal{L}_{g}(\theta,\mathcal{A})
=
-
\sum_{k=1}^{K}
\frac{S_k^{\mathrm{intra}}}
{S_k^{\mathrm{inter}}+\epsilon},
\label{eq:rewrite_group_loss}
\end{equation}
where \(\theta\) denotes the feature extractor parameters. Minimizing \(\mathcal{L}_g\) maximizes the intra-to-inter group similarity ratio. CIR does not assign semantic labels to groups; concept supervision provides the semantic target.

\paragraph{Training objective.}
We add CIR to the standard CBM objective:
\begin{equation}
\mathcal{L}
=
\mathcal{L}_{\mathrm{CBM}}(\theta,\psi)
+
\lambda_g\mathcal{L}_{g}(\theta,\mathcal{A}).
\label{eq:rewrite_full_objective}
\end{equation}
The task and concept losses determine what the model predicts, while CIR constrains how filters organize to support those predictions. CIR therefore shapes representation structure without replacing concept supervision. Full definitions of the partition and all diagnostic operators are provided in Appendix~\ref{app:complete_integrity_definitions}.
 
\section{Experiments}
\label{sec:experiments_rewrite}

The experiments test whether concept integrity provides information beyond concept and task accuracy, then examine three behavioral evaluations in which representation quality matters. Background masking evaluates sensitivity to nuisance context, concept intervention evaluates bottleneck controllability, and limited concept labels evaluate annotation efficiency. Throughout, we interpret concept and task accuracy jointly rather than as a single ranking, and use retention as the core metric under label scarcity.

\subsection{Experimental Setup}
\label{sec:rewrite_experiment_setup}

\paragraph{Datasets.}
We evaluate on three concept-annotated visual recognition benchmarks with different granularities. CUB is a fine-grained bird recognition benchmark with attribute annotations commonly used in CBM evaluation~\cite{kohConceptBottleneckModels2020a}. CelebA provides face-attribute concepts~\cite{liuDeepLearningFace2015}. AwA2 provides animal-attribute concepts~\cite{xianZeroShotLearningComprehensive2020}. GC, CC, and CIS are reported at the fixed analysis resolution $K=16$ on all three datasets. The LDCBM group-count sensitivity (Appendix~\ref{app:k_sensitivity}), training-order control (Appendix~\ref{app:training_order}), and label-scarce experiments use CUB. Background masking and concept intervention are evaluated on CUB, AwA2, and CelebA.

\paragraph{Baselines.}
We compare against two concept-based baselines. \textbf{CBM} is the standard supervised concept bottleneck model~\cite{kohConceptBottleneckModels2020a}. \textbf{CEM} represents each concept with an embedding rather than only a scalar bottleneck value~\cite{zarlengaConceptEmbeddingModels2022a}, and tests whether a richer bottleneck automatically yields intact concept support. We refer to the standard CBM trained with CIR as the \textbf{latent decoupled concept bottleneck model (LDCBM)}; the name reflects the group-level latent structure that CIR introduces, not an additional architectural component.

\paragraph{Metrics.}
We report concept accuracy, task accuracy, and representation integrity. Concept and task accuracy are interpreted jointly: a useful CBM must preserve both concept fidelity and downstream prediction, so a high value on one predictive metric does not compensate for a collapsed value on the other. GC, CC, and CIS are computed as defined in Sec.~\ref{sec:rewrite_metrics}. These integrity metrics provide representation-level information rather than replacing either predictive metric. Under label scarcity, the core metric is task retention, $\operatorname{Retention}=A_{\mathrm{limited}}/A_{\mathrm{full}}$, where $A_{\mathrm{limited}}$ is task accuracy with limited concept labels and $A_{\mathrm{full}}$ is the corresponding full-label task accuracy; concept accuracy is reported as secondary context.

\paragraph{Implementation details.}
All models use ResNet-18 as the backbone. GC, CC, CIS, and CIR operate on the pooled activation of layer3. The CIR objective follows Eq.~\ref{eq:rewrite_full_objective} with a single weight $\lambda_g$; the reported LDCBM configuration uses $\lambda_g=0.04$ (the two values $\lambda_1, \lambda_2$ in earlier logs correspond to the intra-group and inter-group components of the same objective and are consolidated here into $\lambda_g$). Other LDCBM hyperparameters are learning rate $4\times 10^{-4}$, weight decay $3.2\times 10^{-6}$, concept-loss weight $\alpha=0.8$, classification-loss weight $\beta=0.4$, group center number $K=16$, and temperature $t=5$. The selected configuration does not use annealing. CBM and CEM are retrained under their own validation-selected hyperparameters for a fair comparison.

\section{Results}
\label{sec:results_rewrite}

We organize the results around the role of concept integrity in CBMs. We first test whether GC, CC, and CIS provide representation-level information beyond concept and task accuracy. We then examine whether the resulting integrity profiles are associated with behavior under background masking and concept intervention. Finally, we ask whether organized concept support also improves the efficiency with which models use limited concept labels.
\subsection{Predictive Accuracy Does Not Determine Concept Integrity}
\label{sec:rewrite_main_integrity}

\begin{table}[t]
\caption{Cross-dataset comparison of predictive accuracy and group-level concept integrity at $K=16$. Concept and task accuracy are reported jointly. CIS is the geometric mean of GC and CC. All entries are computed from one training run.}
\label{tab:rewrite_main_results}
\begin{center}
\small
\setlength{\tabcolsep}{4.6pt}
\renewcommand{\arraystretch}{1.08}
\begin{tabular}{@{}llccccc@{}}
\toprule
\textbf{Dataset} & \textbf{Model}
& \multicolumn{3}{c}{\textbf{Integrity metrics} $\uparrow$}
& \multicolumn{2}{c}{\textbf{Predictive metrics} $\uparrow$} \\
\cmidrule(lr){3-5}\cmidrule(l){6-7}
& & \textbf{GC} & \textbf{CC} & \textbf{CIS} & \textbf{Concept} & \textbf{Task} \\
\midrule
\multirow{3}{*}{CUB}
& CBM   & 0.610 & 0.409 & 0.500 & 0.908 & 0.653 \\
& CEM   & 0.759 & 0.172 & 0.362 & \textbf{0.915} & \textbf{0.688} \\
\rowcolor{black!5}
& \textbf{LDCBM} & \textbf{0.858} & \textbf{0.443} & \textbf{0.617} & 0.906 & 0.687 \\
\midrule
\multirow{3}{*}{CelebA}
& CBM   & \textbf{0.789} & 0.148 & 0.341 & 0.911 & 0.512 \\
& CEM   & 0.779 & 0.173 & 0.367 & 0.913 & \textbf{0.532} \\
\rowcolor{black!5}
& \textbf{LDCBM} & \textbf{0.789} & \textbf{0.206} & \textbf{0.403} & \textbf{0.913} & \textbf{0.532} \\
\midrule
\multirow{3}{*}{AwA2}
& CBM   & 0.681 & 0.243 & 0.406 & 0.933 & 0.754 \\
& CEM   & 0.717 & 0.174 & 0.353 & \textbf{0.941} & \textbf{0.774} \\
\rowcolor{black!5}
& \textbf{LDCBM} & \textbf{0.730} & \textbf{0.264} & \textbf{0.439} & 0.935 & 0.759 \\
\bottomrule
\end{tabular}
\end{center}
\end{table}

Before relating concept integrity to downstream behavior, we first ask whether it provides information beyond concept and task accuracy without relying on the CIR-trained model. We therefore compare CBM and CEM, two independently motivated concept-based architectures. On CUB and AwA2, CEM achieves higher concept and task accuracy than CBM but lower CIS. Their predictive ranking therefore does not determine their integrity ranking. CelebA shows a similar separation under relatively small predictive differences. These comparisons demonstrate that GC, CC, and CIS capture variation in concept-support organization that remains hidden under output-level evaluation.

We next use LDCBM as a controlled interventional comparison in 
Table~\ref{tab:rewrite_main_results}. Relative to the standard CBM, applying CIR changes the integrity profile while maintaining comparable concept and task accuracy. The effect is particularly clear on CUB, where GC increases from 0.610 to 0.858 while predictive performance remains nearly unchanged. Because CIR modifies filter similarity rather than directly optimizing GC or CC, this comparison shows that the diagnosed representation structure is responsive to a targeted training intervention. Accordingly, the CBM–CEM comparison establishes metric non-redundancy, whereas the CBM–LDCBM comparison establishes that representation integrity is an actionable property.

This separation is not simply an inverse reflection of accuracy: on CelebA, nearly identical predictive performance coexists with distinct CIS. We report \(K=16\) as the primary analysis resolution and defer the coherence--coverage sensitivity across group counts to Appendix~\ref{app:k_sensitivity}. More notably, LDCBM exhibits a counter-intuitive result: CIR improves representation integrity without imposing a predictive penalty. Appendix~\ref{app:cir_accuracy_explanation} provides a detailed analysis of this effect.

\subsection{GC and CC Diagnose Distinct Representation Profiles}
\label{sec:rewrite_integrity_profiles}

Having shown that integrity separates from accuracy, we next interpret what GC and CC add to CIS. These two quantities should not be read as independent objectives that every model must maximize on every dataset. Instead, they decompose concept integrity into two components: semantic coherence within a group and non-fragmented support for a concept.

\begin{figure}[t]
\centering
\includegraphics[scale = 0.7]{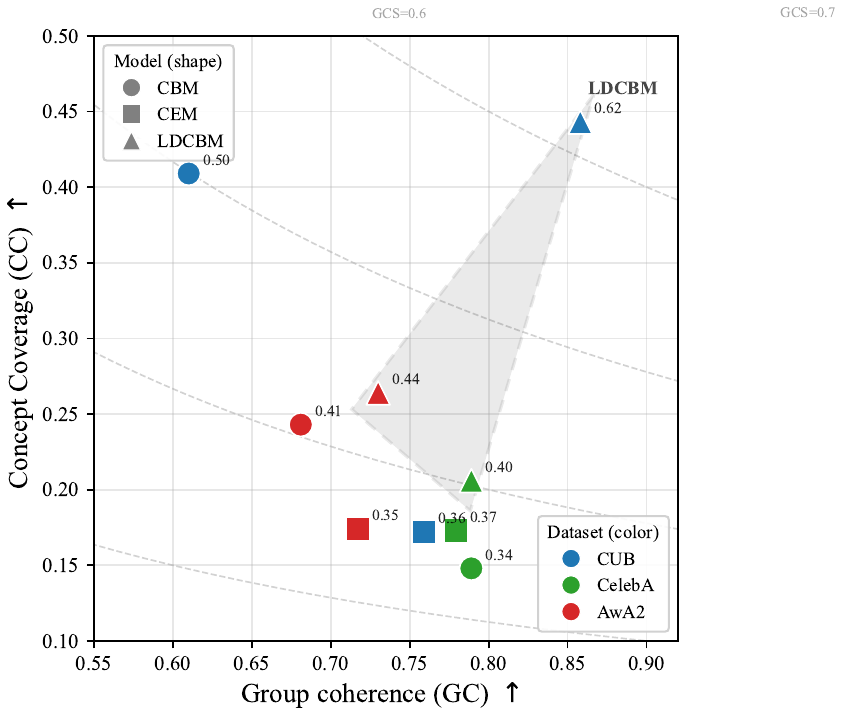}
\caption{Cross-dataset GC--CC profiles at $K=16$. Marker shape denotes the model family, color denotes the dataset, and dashed contours show constant CIS. Labels identify the LDCBM point for each dataset.}
\label{fig:rewrite_cc_gc_scatter}
\end{figure}

Table~\ref{tab:rewrite_main_results} and Fig.~\ref{fig:rewrite_cc_gc_scatter} show this decomposition. CEM's relatively high GC with low CC indicates coherent yet fragmented support, whereas CBM can recover more coverage with weaker coherence. Across all three datasets, the selected LDCBM points lie on higher CIS contours than both baselines. This supports a balanced-profile claim, not coordinate-wise dominance in the GC--CC plane.

The CUB points for CBM and LDCBM sit apart from the denser CelebA and AwA2 cluster. This is consistent with CUB's many fine-grained bird attributes: a richer concept set gives GC and CC more specific targets for dominant-concept and primary-group assignment. The CUB separation therefore reflects dataset granularity as well as learned representation.

Together, GC and CC reveal whether a model's integrity is limited by semantic incoherence, fragmented concept support, or both. The next section examines whether these representation profiles are reflected in model behavior under background masking and concept intervention.

\subsection{Background Masking Reveals Lower Concept and Task Sensitivity for LDCBM}
\label{sec:rewrite_behavior}

\begin{table}[t]
\caption{Background-masking robustness across datasets. Original and masked concept and task accuracies are reported with the relative drop rate. Lower drop indicates greater robustness. Drop rates are computed from one run per model and dataset.}
\label{tab:rewrite_mask}
\begin{center}
\small
\setlength{\tabcolsep}{2.4pt}
\renewcommand{\arraystretch}{1.08}
\begin{tabular}{@{}llccccccc@{}}
\toprule
\textbf{Dataset} & \textbf{Model}
& \textbf{CC} $\uparrow$
& \multicolumn{3}{c}{\textbf{Concept accuracy}}
& \multicolumn{3}{c}{\textbf{Task accuracy}} \\
\cmidrule(lr){4-6}\cmidrule(l){7-9}
& & & \textbf{Orig.} $\uparrow$ & \textbf{Mask.} $\uparrow$ & \textbf{Drop} $\downarrow$
& \textbf{Orig.} $\uparrow$ & \textbf{Mask.} $\uparrow$ & \textbf{Drop} $\downarrow$ \\
\midrule
\multirow{3}{*}{CUB}
& CBM & 0.409 & 0.908 & 0.853 & 6.09\% & 0.653 & 0.360 & 44.87\% \\
& CEM & 0.172 & 0.915 & 0.861 & 5.89\% & 0.688 & 0.354 & 48.51\% \\
\rowcolor{black!5}
& \textbf{LDCBM} & \textbf{0.443} & 0.906 & 0.857 & \textbf{5.45\%} & 0.687 & 0.402 & \textbf{41.50\%} \\
\midrule
\multirow{3}{*}{AwA2}
& CBM & 0.243 & 0.933 & 0.902 & 3.33\% & 0.754 & 0.452 & 40.00\% \\
& CEM & 0.174 & 0.941 & 0.912 & 3.08\% & 0.774 & 0.371 & 52.11\% \\
\rowcolor{black!5}
& \textbf{LDCBM} & \textbf{0.264} & 0.935 & 0.909 & \textbf{2.75\%} & 0.759 & 0.542 & \textbf{28.57\%} \\
\midrule
\multirow{3}{*}{CelebA}
& CBM & 0.148 & 0.911 & 0.875 & 3.93\% & 0.512 & 0.300 & 41.32\% \\
& CEM & 0.173 & 0.913 & 0.883 & 3.34\% & 0.532 & 0.268 & 49.71\% \\
\rowcolor{black!5}
& \textbf{LDCBM} & \textbf{0.206} & 0.913 & 0.888 & \textbf{2.75\%} & 0.532 & 0.370 & \textbf{30.40\%} \\
\bottomrule
\end{tabular}
\end{center}
\end{table}

Background masking turns integrity into a stress test.
If concept support relies on background cues unrelated to the target concept, removing the background should hurt both concept prediction and downstream task prediction.

Across all three datasets, LDCBM shows the smallest drop in both concept accuracy and task accuracy (Table~\ref{tab:rewrite_mask}). Concept drops are modest but consistent: LDCBM drops by 5.45\% on CUB, 2.75\% on AwA2, and 2.75\% on CelebA. The corresponding task drops are larger, but the ordering is again consistent: LDCBM drops by 41.50\%, 28.57\%, and 30.40\%, respectively. These results show that the masking effect is not only a downstream-classifier effect; the concept predictions themselves are also less sensitive to background removal under LDCBM.

The relationship between integrity components and masking behavior is output-specific rather than governed by a single metric. For task accuracy, the drop ranking on CUB and AwA2 follows CC: LDCBM has the highest CC and the smallest task drop, while CEM has the lowest CC and the largest task drop. This resolves the earlier ambiguity on CUB, where CEM's low CC was not reflected by the previous task-drop value. This pattern is consistent with the interpretation that concentrated concept evidence leaves fewer task-relevant filters exposed to spurious background correlations.

A natural concern is that the drop differences simply reflect differences in original accuracy. The updated table weakens this explanation. On AwA2, CEM's original task accuracy is only 0.015 higher than LDCBM's, yet its task drop is 23.54 percentage points larger. On CelebA, CEM and LDCBM have the same original task accuracy, but CEM drops by 49.71\% while LDCBM drops by 30.40\%. The masking results therefore support an association between more organized concept support and lower background sensitivity, while leaving the exact component-level mechanism dataset-dependent.

Background masking perturbs the input; we next test whether stronger integrity also makes the bottleneck itself more controllable under intervention.

\subsection{Stronger Integrity Is Associated with More Responsive Concept Interventions}
\label{sec:rewrite_intervention}

\begin{figure}[tbp]
\centering
\includegraphics[width=0.90\textwidth]{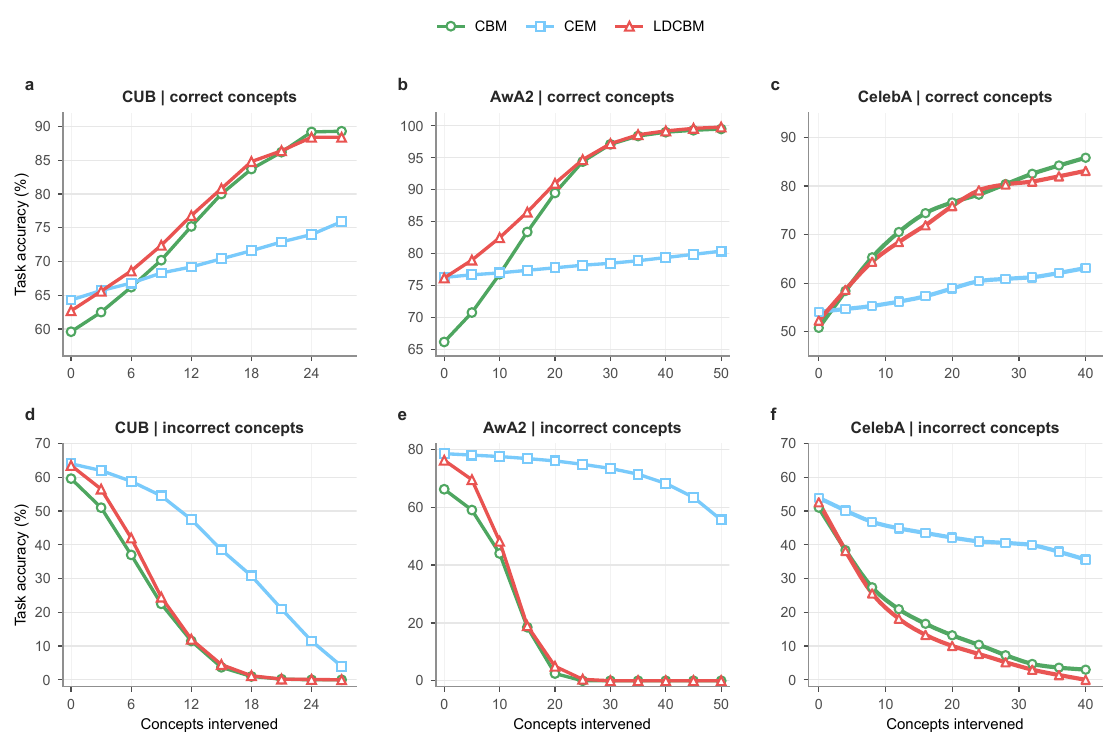}
\caption{Concept intervention curves on CUB, AwA2, and CelebA. Correct interventions replace predicted concepts with ground-truth values; incorrect interventions corrupt them. Each curve is computed from one training run.}
\label{fig:rewrite_intervention}
\end{figure}

Intervention tests the other side of the bottleneck: whether changing concept values actually changes the task prediction. 
A model that bypasses the bottleneck should be less responsive to concept correction or corruption.

Higher CIS is associated with a bottleneck that responds more strongly to concept-level interventions in both directions: gains under correct intervention and losses under incorrect intervention. Intervention requires both that groups be semantically aligned (so intervening on concept $i$ changes the right filters) and that evidence be concentrated (so the intervention touches most of the concept's evidence); GC or CC alone cannot guarantee both, so CIS, as their geometric mean, is the appropriate aggregate integrity score. On CUB, LDCBM's task-accuracy gain from predicted to ground-truth concepts approaches CBM's, while CEM's gain is smaller (Fig.~\ref{fig:rewrite_intervention}). Under full incorrect intervention, LDCBM's task-accuracy drop is larger than CEM's. The patterns on AwA2 and CelebA are qualitatively similar: LDCBM's correct-intervention response approaches CBM's, and its incorrect-intervention drop exceeds CEM's.

Under the method's hypothesis, fewer non-concept channels to the downstream classifier mean that task prediction depends more on the intervenable concept values. Correct interventions then help more, because task-relevant signal is restored, and incorrect interventions hurt more, because task-relevant signal is disrupted. A rival explanation is that LDCBM's larger intervention response reflects a larger bottleneck capacity rather than reduced bypassing. The incorrect-intervention curve argues against this: if bypassing were unchanged, incorrect interventions would not produce a larger drop. The symmetric bidirectional response is consistent with reduced bypassing, not with a mere capacity increase. We describe this as responsiveness rather than leakage: the result is consistent with reduced bypassing of the bottleneck, but should not be interpreted as a direct causal measurement of information leakage.

Background masking and intervention probe behavior under perturbation; we next test whether the same representation structure also helps when concept supervision itself is scarce.

\subsection{Task Retention under Scarce Concept Labels Tracks GC}
\label{sec:rewrite_ssl}

The preceding experiments show that integrity profiles provide information beyond accuracy and are associated with robustness-related behavior. This motivates a further question about annotation efficiency: if a model learns more coherent and less fragmented concept support, can it use limited concept supervision more effectively? We study this question under concept-label scarcity, where the bottleneck dimensionality is unchanged but only a fraction of concept labels is observed during training.

\begin{table}[H]
\caption{Concept-label efficiency on CUB. Full-label columns use all concept labels; 20\% and 10\% columns use partial concept supervision. Retention is defined as $A_{\mathrm{limited}}/A_{\mathrm{full}}$. Each label-fraction setting is reported from one training run.}
\label{tab:rewrite_ssl}
\begin{center}
\scriptsize
\setlength{\tabcolsep}{3.8pt}
\renewcommand{\arraystretch}{1.08}
\begin{tabular}{@{}lccccccccc@{}}
\toprule
\textbf{Model}
& \multicolumn{1}{c}{\textbf{Integrity (100\% labels)}}
& \multicolumn{2}{c}{\textbf{100\% labels} $\uparrow$}
& \multicolumn{3}{c}{\textbf{20\% labels} $\uparrow$}
& \multicolumn{3}{c}{\textbf{10\% labels} $\uparrow$} \\
\cmidrule(lr){2-2}\cmidrule(lr){3-4}\cmidrule(lr){5-7}\cmidrule(l){8-10}
& \textbf{GC} $\uparrow$ & \textbf{Concept} & \textbf{Task}
& \textbf{Concept} & \textbf{Task} & \textbf{Retention}
& \textbf{Concept} & \textbf{Task} & \textbf{Retention} \\
\midrule
CBM & 0.610 & 0.908 & 0.653 & 0.896 & 0.241 & 37\% & 0.855 & 0.104 & 16\% \\
CEM & 0.759 & 0.915 & 0.688 & 0.901 & 0.607 & 88\% & 0.901 & 0.594 & 86\% \\
\rowcolor{black!5}
\textbf{LDCBM} & \textbf{0.858} & 0.906 & 0.687 & 0.885 & 0.637 & \textbf{93\%} & 0.883 & 0.635 & \textbf{92\%} \\
\bottomrule
\end{tabular}
\end{center}
\end{table}

\begin{figure}[tbp]
\centering
\includegraphics[width=0.88\textwidth]{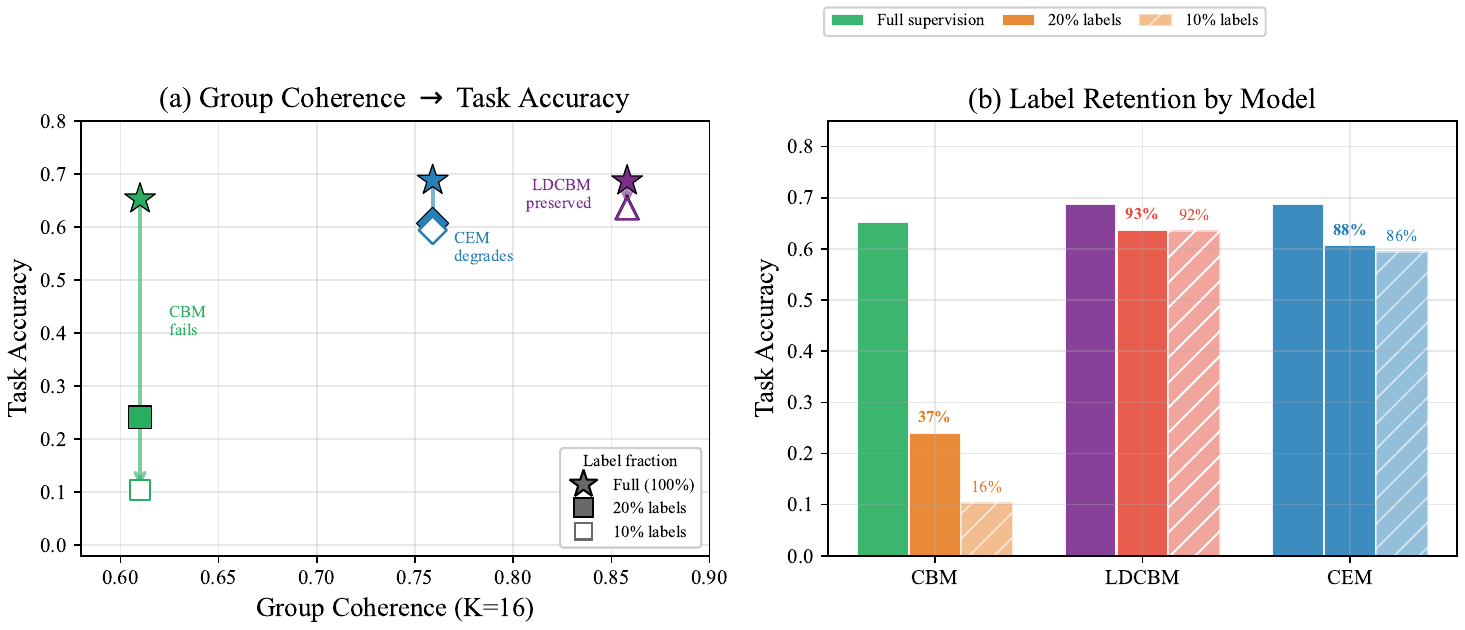}
\caption{Concept-label efficiency on CUB. The left panel places task accuracy against GC under full and limited concept labels. The right panel reports task retention; Table~\ref{tab:rewrite_ssl} provides the paired concept accuracies.}
\label{fig:rewrite_label_efficiency}
\end{figure}

GC captures the property most directly exposed by this setting: whether filters grouped together agree on the same dominant concept. When GC is high, limited concept supervision is applied to an already coherent support structure. When GC is low, the same supervision must also resolve semantic ambiguity within groups whose filters support different concepts. Thus, GC reflects how much concept-disambiguation work remains for the observed labels, rather than merely how strongly individual filters respond.

The results support this interpretation. LDCBM and CEM, which have the two highest GC values, retain (92)--(93\%) and (86)--(88\%) of their fully supervised task performance, respectively, whereas CBM retains only (16)--(37\%). More importantly, the contrast between CBM and CEM separates semantic coherence from evidence concentration. CBM has higher CC and CIS than CEM, yet substantially lower retention. Concentrating concept-relevant filters in one group is therefore insufficient when that group mixes evidence for different concepts. Under scarce supervision, the predictor must still infer which parts of the mixed group support each concept. By contrast, a semantically coherent group reduces this ambiguity even when its concept evidence is less densely concentrated.

This result also clarifies the complementary roles of the integrity metrics. CIS summarizes whether coherence and coverage are jointly high, but it need not be the most informative quantity for every downstream behavior. Concept-label scarcity primarily stresses semantic assignment, which GC isolates directly. CC instead characterizes fragmentation of concept evidence across groups, a property that can remain important for overall integrity and robustness without determining label-scarce retention. The label-scarcity experiment therefore supports the decomposition underlying our metrics: GC and CC expose distinct representation properties, while CIS summarizes their joint integrity.

This concludes the three practical settings; the consolidated findings and limitations are discussed in Sec.~\ref{sec:conclusion_rewrite}.

\section{Conclusion and Limitations}
\label{sec:conclusion_rewrite}

We have argued that concept supervision in CBMs should be evaluated not only by whether the bottleneck outputs match concept labels, but also by how the internal evidence supporting those outputs is organized. Inspired by neural assemblies, we define concept integrity as a group-level property that combines semantic coherence with non-fragmented support. GC, CC, and CIS operationalize these two requirements, while CIR provides a controlled way to perturb the corresponding group organization during training.

Across three concept-annotated datasets, the results show that concept integrity is not reducible to concept or task accuracy. Models with similar predictive performance can exhibit different integrity profiles, and models with high concept accuracy can still have fragmented or mixed concept support. This supports concept integrity as a complementary representation-level criterion rather than a replacement for standard predictive metrics. Background masking and concept intervention further show that these profiles are associated with robustness-related behavior: models with more organized concept support are less sensitive to nuisance context and more responsive to concept correction or corruption. Finally, concept-label scarcity provides a further stress test. When only a fraction of concept annotations is available, organized concept support helps preserve the concept--task pathway, suggesting that intact representations may improve how efficiently models use limited concept supervision.

The current evidence has several limitations. First, GC, CC, and CIS are label-grounded proxies for concept integrity. They measure how filter groups align with concept labels, but they do not guarantee grounding in the intended semantic evidence. Under weak concept supervision, high measured integrity may reflect shortcut correlations or co-occurring concepts, so the metrics should be interpreted with stress tests such as masking, intervention, and task retention. Second, our claims are associative rather than causal. The experiments show that integrity profiles coincide with robustness-related and label-efficiency behavior, but they do not prove that integrity alone causes each improvement. Third, neural assemblies provide an interpretive lens rather than a biological model of the learned filters. Fourth, the current implementation is built around CNN filter groups; extending the idea to transformer-based CBMs requires redefining the grouping unit. Fifth, the label-efficiency experiments are limited to CUB, and the reported results use a single training run. Broader validation across datasets, architectures, and random seeds is needed to establish stability. Finally, our experiments address scarce concept labels, not corrupted concept labels; robustness to noisy annotations remains outside the present scope.

Overall, the results suggest that concept-labelled bottlenecks are not sufficient for reliable concept-based reasoning. Evaluating and encouraging concept integrity offers a more fine-grained way to study when CBMs provide useful, controllable, and annotation-efficient concept representations.

\FloatBarrier
\bibliographystyle{plainnat}
\bibliography{domin,branch}

@inproceedings{sinhaUnderstandingEnhancingRobustness2023,
  title = {Understanding and Enhancing Robustness of Concept-Based Models},
  author = {Sinha, Sanchit and Huai, Mengdi and Sun, Jianhui and Zhang, Aidong},
  booktitle = {Proceedings of the AAAI Conference on Artificial Intelligence},
  volume = {37},
  pages = {15127--15135},
  year = {2023},
  doi = {10.1609/aaai.v37i12.26765},
  url = {https://ojs.aaai.org/index.php/AAAI/article/view/26765},
  eprint = {2211.16080},
  archiveprefix = {arXiv},
  primaryclass = {cs.LG},
  citationuse = {Significance: This work opens the robustness branch by systematically attacking concept-based models with adversarial input perturbations and proposing adversarial training defenses; Cite when: beginning the discussion of CBM robustness while distinguishing input attacks from noisy concept supervision.}
}

@misc{shenInterpretableCompositionalConvolutional2021,
  title = {Interpretable Compositional Convolutional Neural Networks},
  author = {Shen, Wen and Wei, Zhihua and Huang, Shikun and Zhang, Binbin and Fan, Jiaqi and Zhao, Ping and Zhang, Quanshi},
  year = {2021},
  eprint = {2107.04474},
  archiveprefix = {arXiv},
  primaryclass = {cs.CV},
  doi = {10.48550/arXiv.2107.04474},
  url = {https://arxiv.org/abs/2107.04474},
  citationuse = {Significance: This work develops compositional CNN constraints that group filters into coherent visual patterns and provides the grouping lineage used by CPG; Cite when: introducing grouped visual representations or explaining the origin of the filter-similarity partition.}
}

@article{knabWhatsBottleSurvey2026,
  title = {What's in the Bottle? A Survey and Roadmap of Concept Bottleneck Models},
  author = {Knab, Patrick and Steinmann, David and Bartelt, Christian and Kersting, Kristian and Schiele, Bernt and Seidl, Thomas and Schlegel, Udo and Stammer, Wolfgang},
  journal = {Transactions on Machine Learning Research},
  year = {2026},
  month = may,
  url = {https://openreview.net/forum?id=IF5vnqxBEW},
  citationuse = {Significance: This survey organizes more than one hundred CBM studies around input, concept, output, and training modules while identifying grounding, leakage, reliability, and validation as unresolved challenges; Cite when: introducing the rapid expansion and fragmented design space of CBMs or motivating representation-level evaluation beyond predictive accuracy.}
}

@inproceedings{tapliRethinkingConceptBottleneck2026,
  title = {Rethinking Concept Bottleneck Models: From Pitfalls to Solutions},
  author = {Tapli, Merve and Bouniot, Quentin and Stammer, Wolfgang and Akata, Zeynep and Akbas, Emre},
  booktitle = {Proceedings of the IEEE/CVF Conference on Computer Vision and Pattern Recognition},
  pages = {9901--9910},
  year = {2026},
  month = jun,
  url = {https://openaccess.thecvf.com/content/CVPR2026/html/Tapli_Rethinking_Concept_Bottleneck_Models_From_Pitfalls_to_Solutions_CVPR_2026_paper.html},
  eprint = {2603.05629},
  archiveprefix = {arXiv},
  primaryclass = {cs.CV},
  citationuse = {Significance: This work shows that high accuracy can persist with irrelevant concepts and that linear CBM pipelines may bypass the intended bottleneck; Cite when: arguing that output accuracy alone cannot establish concept faithfulness or when motivating model-aware CBM evaluation.}
}

@inproceedings{kohConceptBottleneckModels2020a,
  title = {Concept Bottleneck Models},
  author = {Koh, Pang Wei and Nguyen, Thao and Tang, Yew Siang and Mussmann, Stephen and Pierson, Emma and Kim, Been and Liang, Percy},
  booktitle = {Proceedings of the 37th International Conference on Machine Learning},
  series = {Proceedings of Machine Learning Research},
  volume = {119},
  pages = {5338--5348},
  publisher = {PMLR},
  year = {2020},
  month = jul,
  url = {https://proceedings.mlr.press/v119/koh20a.html},
  citationuse = {Significance: This paper establishes the standard two-stage CBM formulation and concept intervention paradigm; Cite when: defining CBMs, concept supervision, or the promise of test-time correction through human-understandable concepts.}
}

@inproceedings{zarlengaConceptEmbeddingModels2022a,
  title = {Concept Embedding Models: Beyond the Accuracy-Explainability Trade-Off},
  author = {Espinosa Zarlenga, Mateo and Barbiero, Pietro and Ciravegna, Gabriele and Marra, Giuseppe and Giannini, Francesco and Diligenti, Michelangelo and Shams, Zohreh and Precioso, Frederic and Melacci, Stefano and Weller, Adrian and Li{\`o}, Pietro and Jamnik, Mateja},
  booktitle = {Advances in Neural Information Processing Systems},
  volume = {35},
  pages = {21400--21413},
  year = {2022},
  url = {https://papers.nips.cc/paper_files/paper/2022/hash/867c06823281e506e8059f5c13a57f75-Abstract-Conference.html},
  eprint = {2209.09056},
  archiveprefix = {arXiv},
  primaryclass = {cs.LG},
  citationuse = {Significance: Concept Embedding Models replace scalar bottlenecks with richer concept-specific embeddings to improve task utility and intervention behavior under incomplete supervision; Cite when: reviewing attempts to increase bottleneck expressiveness or positioning CEM as a principal baseline.}
}

@inproceedings{zarlengaRobustMetricsConcept2023,
  title = {Towards Robust Metrics for Concept Representation Evaluation},
  author = {Espinosa Zarlenga, Mateo and Barbiero, Pietro and Shams, Zohreh and Kazhdan, Dmitry and Bhatt, Umang and Weller, Adrian and Jamnik, Mateja},
  booktitle = {Proceedings of the AAAI Conference on Artificial Intelligence},
  volume = {37},
  pages = {11791--11799},
  year = {2023},
  doi = {10.1609/aaai.v37i10.26392},
  url = {https://ojs.aaai.org/index.php/AAAI/article/view/26392},
  citationuse = {Significance: This paper introduces metrics for impurities and purity in concept representations and evaluates their relation to robustness and intervention; Cite when: positioning concept-integrity diagnostics against prior concept-representation evaluation.}
}

@article{holtmaatNeuronalAssemblyFormation2016,
  title = {Functional and Structural Underpinnings of Neuronal Assembly Formation in Learning},
  author = {Holtmaat, Anthony and Caroni, Pico},
  journal = {Nature Neuroscience},
  volume = {19},
  pages = {1553--1562},
  year = {2016},
  doi = {10.1038/nn.4418},
  url = {https://www.nature.com/articles/nn.4418},
  citationuse = {Significance: This review describes neuronal assemblies as populations that encode learned information and support retrieval; Cite when: introducing neural assemblies as an explicitly non-biological analogy for coherent functional groups.}
}

@inproceedings{margeloiuConceptBottleneckLearn2021,
  title = {Do Concept Bottleneck Models Learn as Intended?},
  author = {Margeloiu, Andrei and Ashman, Matthew and Bhatt, Umang and Chen, Yanzhi and Jamnik, Mateja and Weller, Adrian},
  booktitle = {ICLR 2021 Workshop on Responsible AI},
  year = {2021},
  doi = {10.17863/CAM.80941},
  url = {https://www.repository.cam.ac.uk/items/3c62e164-75d2-4a4e-bf94-ae5468b7f50d},
  eprint = {2105.04289},
  archiveprefix = {arXiv},
  primaryclass = {cs.LG},
  citationuse = {Significance: This study finds that learned concept predictions may not correspond to semantically meaningful input evidence even when conventional CBM objectives are satisfied; Cite when: motivating the distinction between concept-labelled outputs and concept-aligned internal support.}
}

@inproceedings{havasiAddressingLeakageConcept2022,
  title = {Addressing Leakage in Concept Bottleneck Models},
  author = {Havasi, Marton and Parbhoo, Sonali and Doshi-Velez, Finale},
  booktitle = {Advances in Neural Information Processing Systems},
  volume = {35},
  pages = {23386--23397},
  year = {2022},
  url = {https://papers.nips.cc/paper/2022/hash/944ecf65a46feb578a43abfd5cddd960-Abstract-Conference.html},
  citationuse = {Significance: This paper formalizes how soft concept predictions can transmit unintended task information and links leakage to incomplete concept sets and limited concept predictors; Cite when: explaining why accurate soft bottlenecks may remain uninterpretable or why intervention behavior can degrade.}
}

@inproceedings{shinCloserLookIntervention2023,
  title = {A Closer Look at the Intervention Procedure of Concept Bottleneck Models},
  author = {Shin, Sungbin and Jo, Yohan and Ahn, Sungsoo and Lee, Namhoon},
  booktitle = {Proceedings of the 40th International Conference on Machine Learning},
  series = {Proceedings of Machine Learning Research},
  volume = {202},
  pages = {31504--31520},
  publisher = {PMLR},
  year = {2023},
  month = jul,
  url = {https://proceedings.mlr.press/v202/shin23a.html},
  citationuse = {Significance: This work demonstrates that intervention effectiveness depends strongly on concept selection, granularity, and operating conditions rather than following automatically from the CBM architecture; Cite when: describing intervention as a central but nontrivial reliability criterion.}
}

@inproceedings{huangConceptTrustworthinessConcept2024a,
  title = {On the Concept Trustworthiness in Concept Bottleneck Models},
  author = {Huang, Qihan and Song, Jie and Hu, Jingwen and Zhang, Haofei and Wang, Yong and Song, Mingli},
  booktitle = {Proceedings of the AAAI Conference on Artificial Intelligence},
  volume = {38},
  pages = {21161--21168},
  year = {2024},
  doi = {10.1609/aaai.v38i19.30109},
  url = {https://ojs.aaai.org/index.php/AAAI/article/view/30109},
  citationuse = {Significance: This paper evaluates whether concept predictions arise from spatially relevant regions and shows that the input-to-concept mapping can remain untrustworthy despite a transparent bottleneck; Cite when: motivating concept grounding or distinguishing output-level concept accuracy from evidence-level trustworthiness.}
}

@article{ramanConceptBottleneckRespect2025,
  title = {Do Concept Bottleneck Models Respect Localities?},
  author = {Raman, Naveen Janaki and Espinosa Zarlenga, Mateo and Heo, Juyeon and Jamnik, Mateja},
  journal = {Transactions on Machine Learning Research},
  year = {2025},
  url = {https://openreview.net/forum?id=4mCkRbUXOf},
  eprint = {2401.01259},
  archiveprefix = {arXiv},
  primaryclass = {cs.LG},
  citationuse = {Significance: This study shows that accurate concept predictors may react to features outside a concept's spatial or semantic locality, making concept explanations fragile; Cite when: arguing that concept accuracy does not ensure that predictions rely on concept-relevant internal evidence.}
}

@misc{liuDeepLearningFace2015,
  title = {Deep Learning Face Attributes in the Wild},
  author = {Liu, Ziwei and Luo, Ping and Wang, Xiaogang and Tang, Xiaoou},
  year = {2015},
  eprint = {1411.7766},
  archiveprefix = {arXiv},
  primaryclass = {cs.CV},
  doi = {10.48550/arXiv.1411.7766},
  url = {https://arxiv.org/abs/1411.7766},
  citationuse = {Significance: This paper introduces the CelebA face-attribute dataset used for evaluating concept prediction at scale; Cite when: describing the CelebA benchmark and its attribute annotations.}
}

@misc{xianZeroShotLearningComprehensive2020,
  title = {Zero-Shot Learning---A Comprehensive Evaluation of the Good, the Bad and the Ugly},
  author = {Xian, Yongqin and Lampert, Christoph H. and Schiele, Bernt and Akata, Zeynep},
  year = {2020},
  eprint = {1707.00600},
  archiveprefix = {arXiv},
  primaryclass = {cs.CV},
  doi = {10.48550/arXiv.1707.00600},
  url = {https://arxiv.org/abs/1707.00600},
  citationuse = {Significance: This work establishes the AwA2 benchmark and its semantic attributes for animal recognition; Cite when: introducing AwA2 as a concept-annotated evaluation dataset.}
}
\appendix
\clearpage
\setlength{\abovedisplayskip}{6pt plus 2pt minus 2pt}
\setlength{\belowdisplayskip}{6pt plus 2pt minus 2pt}
\setlength{\abovedisplayshortskip}{3pt plus 2pt minus 1pt}
\setlength{\belowdisplayshortskip}{6pt plus 2pt minus 2pt}

\section{Group-Count Sensitivity}
\label{app:k_sensitivity}

Is the selected LDCBM grouping sensitive to the number of groups? Figure~\ref{fig:rewrite_cis_k} and Table~\ref{tab:rewrite_k_sensitivity} report GC, CC, and CIS for LDCBM across $K\in\{8,16,32,64\}$ on CUB. We do not sweep CBM or CEM here because $K$ is a CIR grouping hyperparameter, not a training parameter for those baselines. Increasing $K$ changes the granularity of the learned groups. With more groups, each group contains fewer filters, so a small set of concept-selective activated features can dominate the group assignment and make high GC easier to obtain. The same granularity also fragments the filters relevant to a given concept across multiple groups. The primary group for that concept then captures a smaller fraction of its relevant-filter set, reducing CC. CIS therefore decreases as groups become too fine, from 0.693 at $K=8$ to 0.383 at $K=64$. The main experiments use $K=16$ as a fixed operating point for LDCBM and use the same resolution only as a diagnostic setting for CBM and CEM.

\par\smallskip
{\centering
\includegraphics[width=0.95\columnwidth]{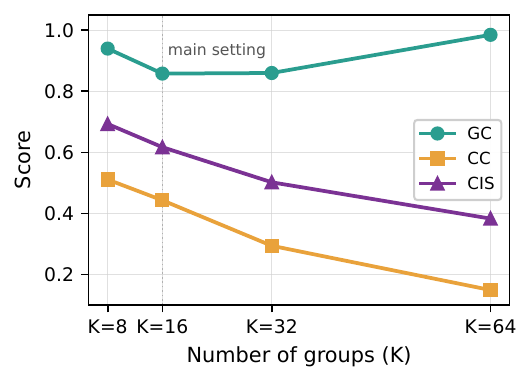}
\refstepcounter{figure}\label{fig:rewrite_cis_k}
\vspace{1pt}
\begin{minipage}{0.95\columnwidth}
\small Figure~\thefigure: LDCBM group-count sensitivity on CUB. Changes across $K$ reflect the balance between semantic coherence and non-fragmentation.
\end{minipage}
\par}
\smallskip

\par\smallskip
\noindent\begin{minipage}{\columnwidth}
\refstepcounter{table}\label{tab:rewrite_k_sensitivity}
\small Table~\thetable: LDCBM group-count sensitivity on CUB. GC, CC, and CIS are reported across four CIR group counts. The shaded row indicates the main-text setting. Each $K$ setting uses one LDCBM training run.
\par\smallskip
\centering
\small
\setlength{\tabcolsep}{5.0pt}
\renewcommand{\arraystretch}{1.08}
\begin{tabular}{@{}lccc@{}}
\toprule
\textbf{$K$} & \textbf{GC} $\uparrow$ & \textbf{CC} $\uparrow$ & \textbf{CIS} $\uparrow$ \\
\midrule
8  & 0.940 & \textbf{0.511} & \textbf{0.693} \\
\rowcolor{black!5}
16 & 0.858 & 0.443 & 0.617 \\
32 & 0.860 & 0.294 & 0.502 \\
64 & \textbf{0.985} & 0.149 & 0.383 \\
\bottomrule
\end{tabular}
\end{minipage}
\vspace{-4pt}

\section{Qualitative Representation Analysis}
\label{sec:rewrite_qualitative}

\begin{figure}[t]
\centering
\includegraphics[width=0.82\textwidth]{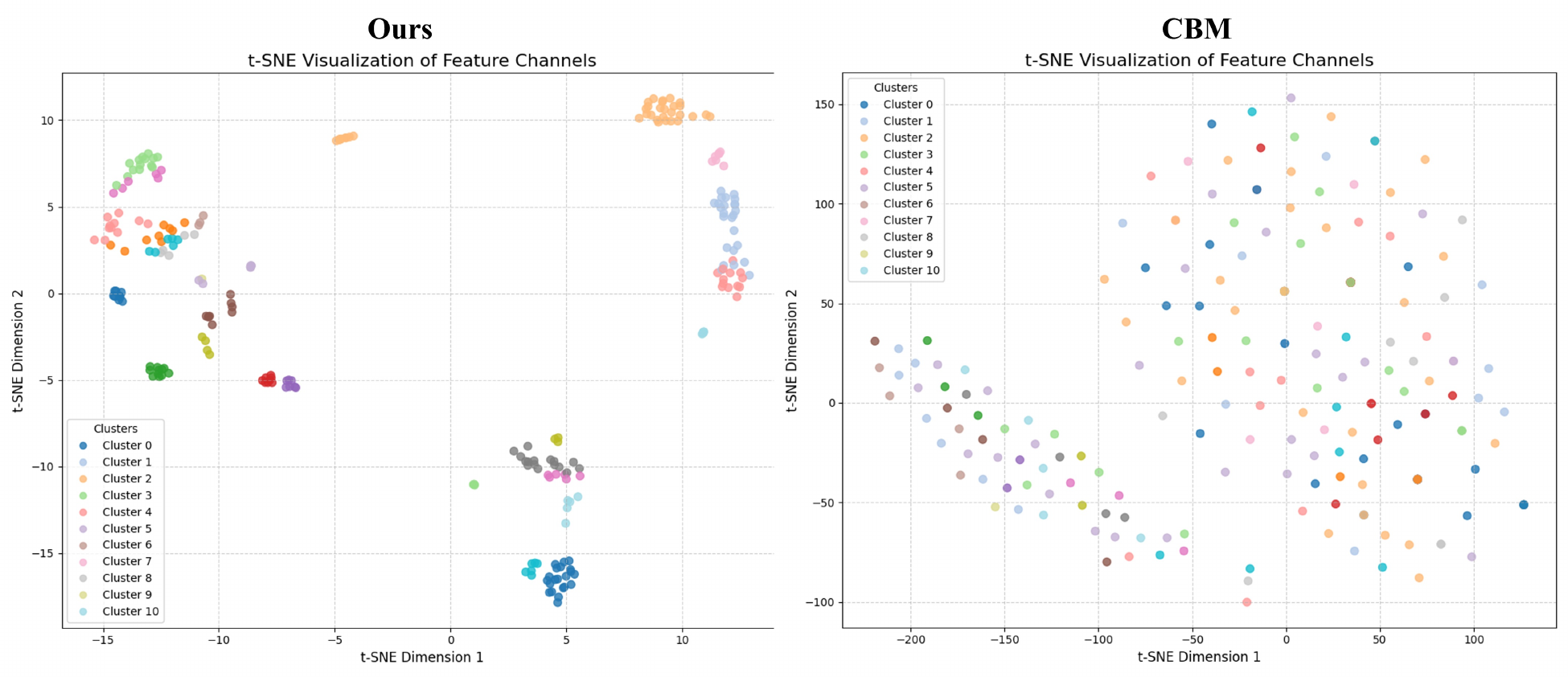}
\caption{t-SNE visualization of feature distributions before the concept layer. LDCBM forms more separated feature clusters than vanilla CBM, consistent with the group-organization objective used by CIR.}
\label{fig:rewrite_tsne}
\end{figure}

The t-SNE visualization provides qualitative support that LDCBM forms more separated feature distributions than vanilla CBM. However, cluster separation alone does not establish concept integrity; it is used only as a visual complement to the quantitative GC, CC, and CIS diagnostics.

\subsection{Concept-Attribution Case Study}
\label{app:concept_attribution_case}

\begin{figure}[p]
\centering
\includegraphics[width=\textwidth,height=0.82\textheight,keepaspectratio]{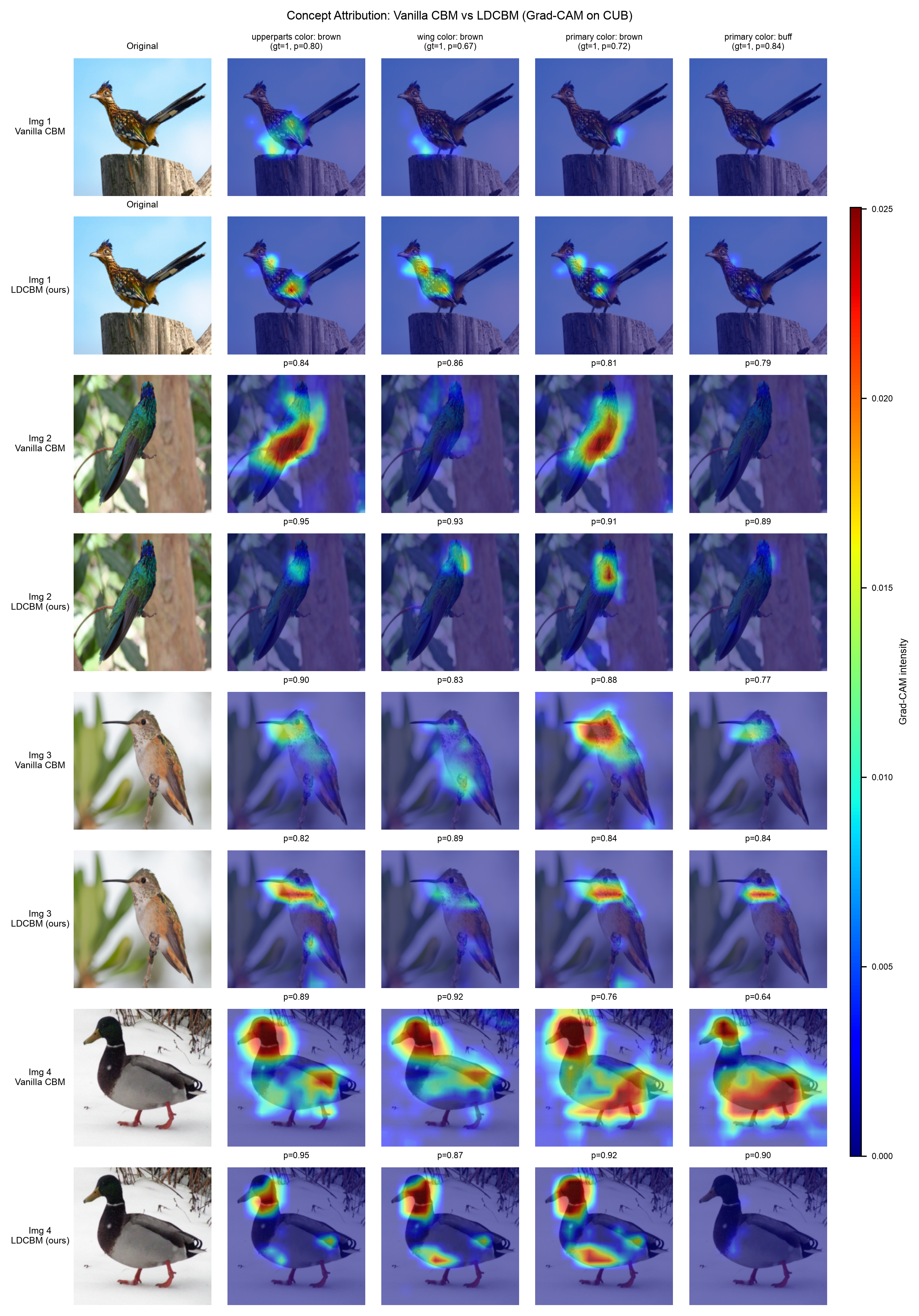}
\caption{Qualitative concept-attribution case study on CUB. Rows compare vanilla CBM and LDCBM on the same images using Grad-CAM maps for selected concepts; the numbers above panels denote predicted concept probabilities. Compared with vanilla CBM, LDCBM places less attribution on background or adjacent non-concept regions in these examples, consistent with the reduced concept-shift behavior observed for the stronger-integrity LDCBM variant.}
\label{fig:concept_attribution_case}
\end{figure}

Figure~\ref{fig:concept_attribution_case} provides a qualitative case study of concept attribution on CUB. For the same image and selected concepts, the vanilla CBM often assigns salient attribution to background or adjacent non-concept regions, whereas LDCBM tends to concentrate attribution more tightly on visible bird parts in these examples. This pattern is consistent with the main quantitative evidence that higher concept integrity is associated with reduced concept-shift behavior. The visualization is used as an interpretive case study rather than as an additional metric.

\FloatBarrier

\section{Training-Schedule Control}
\label{app:training_order}

\par\smallskip
\noindent\begin{minipage}{\columnwidth}
\refstepcounter{table}\label{tab:rewrite_annealing}
Table~\thetable: Training-schedule control on CUB. CIS is reported for LDCBM at $K=16$, with $\Delta$CIS computed relative to no annealing. Each strategy is evaluated with one run.
\par\smallskip
\centering
\small
\setlength{\tabcolsep}{5.5pt}
\renewcommand{\arraystretch}{1.08}
\begin{tabular}{@{}lcc@{}}
\toprule
\textbf{Training strategy} & \textbf{CIS} $\uparrow$ & \textbf{$\Delta$CIS} \\
\midrule
No annealing & \textbf{0.617} & -- \\
Annealing & 0.370 & -0.247 \\
Two-phase training & 0.360 & -0.257 \\
\bottomrule
\end{tabular}
\end{minipage}

Can concept-first scheduling substitute for CIR? Table~\ref{tab:rewrite_annealing} reports CIS under three training schedules on CUB with LDCBM at $K=16$. Annealing reduces CIS from 0.617 to 0.37, and two-phase training reduces it to 0.36. Concept-first scheduling does not reproduce the group-level organization that CIR encourages. The selected configuration therefore used CIR without annealing.

\FloatBarrier
\section{Complete Definitions of Concept Integrity Metrics and CIR}
\label{app:complete_integrity_definitions}

This appendix gives the implementation-level definitions used for GC, CC, CIS, and Concept Integrity Regularization (CIR). The main text introduces the metrics conceptually. Here we specify the activation matrix, filter groups, adaptive relevant-filter sets, aggregation rules, and numerical safeguards used to compute them.

\subsection{Activation Vectors and Filter Groups}

All diagnostics and CIR are computed from the pooled activations of the same convolutional layer. In our experiments this layer is the ResNet-18 layer3 output. For a sample \(x^{(n)}\) and filter \(u\in\Omega=\{1,\ldots,|\Omega|\}\), the scalar activation is
\begin{equation*}
a_u^{(n)}
=
\operatorname{GAP}\!\left(z_u(x^{(n)})\right),
\end{equation*}
where \(z_u(x^{(n)})\) denotes the spatial activation map of filter \(u\). The activation vector of filter \(u\), the activation matrix, and the concept matrix are
\begin{equation*}
\begin{aligned}
\mathbf{a}_u
&=
[a_u^{(1)},\ldots,a_u^{(N)}]^\top,\\
\mathbf{A}
&=
[\mathbf{a}_1,\ldots,\mathbf{a}_{|\Omega|}],\\
\mathbf{C}
&=
[\mathbf{c}^{(1)},\ldots,\mathbf{c}^{(N)}]^\top .
\end{aligned}
\end{equation*}
The matrix \(\mathbf{C}\in\{0,1\}^{N\times M}\) contains the \(M\) binary concept annotations.

Filter groups are obtained by spectral clustering on the filter-similarity matrix defined below. For a fixed dataset and group count \(K\), the resulting partition in Eq.~\ref{eq:app_group_partition} is stored and reused for the corresponding metric computation:
\begin{equation}
\begin{aligned}
\mathcal{A}
&=
\{\mathcal{A}_1,\ldots,\mathcal{A}_K\},\\
\Omega
&=
\bigcup_{k=1}^{K}\mathcal{A}_k,\\
\mathcal{A}_i\cap\mathcal{A}_j
&=
\varnothing\quad\text{for }i\neq j.
\end{aligned}
\label{eq:app_group_partition}
\end{equation}
The same partition \(\mathcal{A}\) is used when computing GC, CC, CIS, and CIR for a given setting. This avoids comparing metrics that are computed from different grouping resolutions.

\subsection{Filter--Concept Selectivity}

The selectivity of filter \(u\) to concept \(i\), defined in Eq.~\ref{eq:app_selectivity}, is the absolute Pearson correlation between the filter activation vector and the concept annotation vector:
\begin{equation}
q_i(u)
=
\left|
\operatorname{corr}
(\mathbf{a}_u,\mathbf{C}_{:,i})
\right|,
\qquad
Q_{u,i}=q_i(u).
\label{eq:app_selectivity}
\end{equation}
The absolute value makes the score sensitive to both positive and negative concept-selective responses. Constant activation or concept vectors are assigned zero correlation. Unless otherwise stated, all reported GC, CC, and CIS values use a small numerical constant \(\epsilon=10^{-8}\) in denominators.

\subsection{Group Coherence}

Each group is assigned to the concept with the largest mean selectivity among its filters, as in Eq.~\ref{eq:app_group_assignment}:
\begin{equation}
\phi(k)
=
\arg\max_{i\in\{1,\ldots,M\}}
\frac{1}{|\mathcal{A}_k|}
\sum_{u\in\mathcal{A}_k}q_i(u).
\label{eq:app_group_assignment}
\end{equation}
Ties are resolved by the deterministic ordering of concept indices. The local group-coherence score compares each filter's selectivity to the group concept with that filter's strongest concept selectivity:
\begin{equation*}
\operatorname{GC}(k)
=
\frac{1}{|\mathcal{A}_k|}
\sum_{u\in\mathcal{A}_k}
\frac{q_{\phi(k)}(u)}
{\max_{j\in\{1,\ldots,M\}}q_j(u)+\epsilon}.
\end{equation*}
The dataset-level group coherence in Eq.~\ref{eq:app_gc} is the macro-average over groups:
\begin{equation}
\operatorname{GC}
=
\frac{1}{K}\sum_{k=1}^{K}\operatorname{GC}(k).
\label{eq:app_gc}
\end{equation}
High GC indicates that filters in the same group tend to support one dominant concept rather than a mixture of concepts.

\subsection{Adaptive Relevant-Filter Sets and Concept Coverage}

Concept coverage requires a relevant-filter set for each concept. For concept \(i\), we first compute the adaptive selectivity threshold in Eq.~\ref{eq:app_threshold}:
\begin{equation}
\tau_i
=
\begin{cases}
\tau,
& \text{if }\max_{u\in\Omega}q_i(u)>\tau,\\
\operatorname{Quantile}_{0.8}
(\{q_i(u)\}_{u\in\Omega}),
& \text{otherwise},
\end{cases}
\label{eq:app_threshold}
\end{equation}
where \(\tau=0.3\) in our implementation. The relevant-filter set is defined in Eq.~\ref{eq:app_relevant_set}:
\begin{equation}
\mathcal{R}_i
=
\{u\in\Omega\mid q_i(u)>\tau_i\}.
\label{eq:app_relevant_set}
\end{equation}
If this strict threshold produces an empty set, \(\mathcal{R}_i\) is set to the filter with the largest \(q_i(u)\). This fallback keeps every concept represented in the macro-average.

The primary group for concept \(i\) is the group with the largest mean selectivity to that concept, as in Eq.~\ref{eq:app_concept_group}:
\begin{equation}
\psi(i)
=
\arg\max_{k\in\{1,\ldots,K\}}
\frac{1}{|\mathcal{A}_k|}
\sum_{u\in\mathcal{A}_k}q_i(u).
\label{eq:app_concept_group}
\end{equation}
Ties are again resolved by deterministic index order. The local concept-coverage score is
\begin{equation*}
\operatorname{CC}(i)
=
\frac{|\mathcal{R}_i\cap\mathcal{A}_{\psi(i)}|}
{|\mathcal{R}_i|+\epsilon}.
\end{equation*}
The reported concept coverage in Eq.~\ref{eq:app_cc} is the macro-average over concepts:
\begin{equation}
\operatorname{CC}
=
\frac{1}{M}\sum_{i=1}^{M}\operatorname{CC}(i).
\label{eq:app_cc}
\end{equation}
High CC means that the filters relevant to a concept are concentrated in that concept's primary group rather than dispersed across many groups.

\subsection{Concept Integrity Score}

Concept integrity score combines semantic coherence and concept coverage through the geometric mean in Eq.~\ref{eq:app_cis}:
\begin{equation}
\operatorname{CIS}
=
\sqrt{\operatorname{GC}\operatorname{CC}}.
\label{eq:app_cis}
\end{equation}
The geometric mean prevents either component from fully compensating for the other. A model therefore receives a high CIS only when concept-supporting groups are both coherent and non-fragmented.

\subsection{Filter Similarity and CIR}

The shifted Pearson similarity between filters \(u\) and \(v\) is defined in Eq.~\ref{eq:app_similarity}:
\begin{equation}
s_{uv}
=
\operatorname{corr}(\mathbf{a}_u,\mathbf{a}_v)+1.
\label{eq:app_similarity}
\end{equation}
The shift maps Pearson correlation from \([-1,1]\) to \([0,2]\). For group \(\mathcal{A}_k\), the within-group and cross-group similarities are defined in Eq.~\ref{eq:app_intra_inter}:
\begin{equation}
\begin{aligned}
S_k^{\mathrm{intra}}
&=
\sum_{u,v\in\mathcal{A}_k}s_{uv},\\
S_k^{\mathrm{inter}}
&=
\sum_{\substack{u\in\mathcal{A}_k\\
v\in\Omega\setminus\mathcal{A}_k}}s_{uv}.
\end{aligned}
\label{eq:app_intra_inter}
\end{equation}
The CIR objective in Eq.~\ref{eq:app_group_loss} is
\begin{equation}
\mathcal{L}_{g}(\theta,\mathcal{A})
=
-
\sum_{k=1}^{K}
\frac{S_k^{\mathrm{intra}}}
{S_k^{\mathrm{inter}}+\epsilon}.
\label{eq:app_group_loss}
\end{equation}
Minimizing \(\mathcal{L}_g\) maximizes the intra-to-inter group similarity ratio. CIR shapes how filters organize, while concept supervision provides the semantic target.

The complete training objective in Eq.~\ref{eq:app_full_objective} is
\begin{equation}
\mathcal{L}
=
\mathcal{L}_{\mathrm{CBM}}(\theta,\psi)
+
\lambda_g\mathcal{L}_{g}(\theta,\mathcal{A}).
\label{eq:app_full_objective}
\end{equation}
Here \(\mathcal{L}_{\mathrm{CBM}}\) is the task-plus-concept objective defined in the main text. The selected LDCBM configuration uses \(\lambda_g=0.04\), \(K=16\), and no annealing.

\section{Why CIR Can Improve Accuracy Despite Adding a Constraint}
\label{app:cir_accuracy_explanation}

The main text treats LDCBM as a standard CBM trained with Concept Integrity Regularization (CIR). CIR does not add a new prediction pathway. Instead, it constrains the organization of the shared convolutional filters used to predict concepts. This creates an apparent tension: a constraint should reduce the model's freedom, so why can LDCBM sometimes match or slightly improve concept accuracy and task accuracy relative to CBM?

The key point is that regularization can remove harmful degrees of freedom. If an unconstrained CBM uses nuisance background cues or mixed concept evidence, it can fit the training objective while learning a representation that generalizes poorly. CIR can improve test performance when the reduction in nuisance-sensitive variation is larger than the approximation cost introduced by the constraint.

\subsection{Notation and Objective}

Let \(h \in \mathcal{H}_{\mathrm{CBM}}\) denote a standard CBM, and let \(\mathcal{H}_{\mathrm{CIR}} \subseteq \mathcal{H}_{\mathrm{CBM}}\) denote the subset of CBMs whose filters satisfy the CIR-induced grouping preference. LDCBM is trained by minimizing
\begin{equation}
\mathcal{L}_{\mathrm{LDCBM}}
=
\mathcal{L}_{\mathrm{task}}(f(\hat{\mathbf{c}}),y)
+
\lambda_c \mathcal{L}_{\mathrm{concept}}(\hat{\mathbf{c}},\mathbf{c})
+
\lambda_g \mathcal{L}_{g}.
\label{eq:app_ldcbm_cir_objective}
\end{equation}
This notation matches the current manuscript: CIR is a group-regularization term, not a separate disentanglement loss, and LDCBM does not introduce an explicit residual subspace.

For a model \(h\), let \(R_y(h)\) be the expected task risk and \(R_c(h)\) be the expected concept-prediction risk. Their empirical counterparts on \(N\) training samples are \(\widehat{R}_y(h)\) and \(\widehat{R}_c(h)\). We use a generic complexity term \(\mathfrak{C}(\mathcal{H})\) to denote the effective capacity of a hypothesis class.

\subsection{A Capacity-Bias Explanation}

\paragraph{Lemma 1.}
For a bounded loss and a hypothesis class \(\mathcal{H}\), a standard uniform-convergence form gives, with high probability,
\begin{equation}
R(h)
\leq
\widehat{R}(h)
+
O\!\left(\sqrt{\frac{\mathfrak{C}(\mathcal{H})}{N}}\right).
\label{eq:app_generic_bound}
\end{equation}
Because \(\mathcal{H}_{\mathrm{CIR}} \subseteq \mathcal{H}_{\mathrm{CBM}}\), the effective complexity of the CIR-constrained class can be smaller:
\begin{equation}
\mathfrak{C}(\mathcal{H}_{\mathrm{CIR}})
\leq
\mathfrak{C}(\mathcal{H}_{\mathrm{CBM}}).
\label{eq:app_complexity_order}
\end{equation}
Thus, CIR can reduce the estimation component of the test-risk bound. This does not guarantee better accuracy by itself, because the constraint may also increase empirical or approximation error.

\paragraph{Proposition 1.}
Let \(h_{\mathrm{CBM}}\) be the selected standard CBM and \(h_{\mathrm{LDCBM}}\) be the selected CIR-constrained model. Suppose CIR increases the empirical risk by at most \(\Delta_{\mathrm{app}}\):
\begin{equation}
\widehat{R}(h_{\mathrm{LDCBM}})
-
\widehat{R}(h_{\mathrm{CBM}})
\leq
\Delta_{\mathrm{app}}.
\label{eq:app_approx_cost}
\end{equation}
Suppose also that the reduction in the complexity or nuisance-sensitive component of the bound is at least \(\Delta_{\mathrm{reg}}\):
\begin{equation}
O\!\left(\sqrt{\frac{\mathfrak{C}(\mathcal{H}_{\mathrm{CBM}})}{N}}\right)
-
O\!\left(\sqrt{\frac{\mathfrak{C}(\mathcal{H}_{\mathrm{CIR}})}{N}}\right)
\geq
\Delta_{\mathrm{reg}}.
\label{eq:app_reg_gain}
\end{equation}
If \(\Delta_{\mathrm{reg}} > \Delta_{\mathrm{app}}\), then the resulting test-risk bound for LDCBM is tighter than that for the standard CBM.

\paragraph{Proof.}
Applying Lemma 1 to the two model classes gives
\begin{equation}
R(h_{\mathrm{CBM}})
\leq
\widehat{R}(h_{\mathrm{CBM}})
+
B_{\mathrm{CBM}},
\qquad
R(h_{\mathrm{LDCBM}})
\leq
\widehat{R}(h_{\mathrm{LDCBM}})
+
B_{\mathrm{CIR}},
\end{equation}
where \(B_{\mathrm{CBM}}\) and \(B_{\mathrm{CIR}}\) denote the corresponding complexity terms. By assumption,
\begin{equation}
\widehat{R}(h_{\mathrm{LDCBM}}) \leq
\widehat{R}(h_{\mathrm{CBM}}) + \Delta_{\mathrm{app}},
\qquad
B_{\mathrm{CIR}} \leq B_{\mathrm{CBM}}-\Delta_{\mathrm{reg}}.
\end{equation}
Therefore,
\begin{equation}
R(h_{\mathrm{LDCBM}})
\leq
\widehat{R}(h_{\mathrm{CBM}})
+
B_{\mathrm{CBM}}
-
(\Delta_{\mathrm{reg}}-\Delta_{\mathrm{app}}).
\end{equation}
When \(\Delta_{\mathrm{reg}}>\Delta_{\mathrm{app}}\), the bound for LDCBM is tighter. This establishes a sufficient condition under which a constrained model can have better expected test performance than a less constrained one.

\subsection{Connection to Concept and Task Accuracy}

The same argument applies to both concept risk and task risk. For concept prediction, CIR is useful when the concept-relevant filters can still be represented inside coherent groups, while nuisance-sensitive or semantically mixed filters are discouraged. In that case, the approximation cost for concept prediction is small, but the generalization benefit can be positive.

For task prediction, the downstream classifier receives predicted concepts. If CIR improves the stability or semantic alignment of the concept predictions, task accuracy can also improve. This effect does not require the regularizer to increase raw model capacity. It only requires the regularizer to shift the inductive bias toward features that transfer better from training to test data.

\subsection{Interpretation}

This analysis explains why the observed accuracy pattern is not contradictory. CIR restricts the hypothesis space, so it can reduce the amount of information the model uses. However, the removed information need not be useful concept evidence. If the removed degrees of freedom mainly correspond to nuisance cues, background correlations, or mixed concept support, the constrained model can generalize better.

The statement is intentionally conditional. CIR should not be read as an unconditional guarantee of higher concept accuracy or task accuracy. It predicts improvement only when the regularization benefit exceeds the approximation cost. This matches the empirical framing of the main paper: LDCBM is not presented as a universally higher-capacity model, but as a CIR-constrained CBM whose integrity profile can be associated with better robustness and, in some settings, slightly better predictive accuracy.

\FloatBarrier
\end{document}